\documentclass[conference, 10pt]{IEEEtran}
\IEEEoverridecommandlockouts
\usepackage{cite}
\usepackage{amsmath,amssymb,amsfonts}
\usepackage{algorithmic}
\usepackage{graphicx}
\usepackage{textcomp}
\usepackage[table,xcdraw]{xcolor}
\usepackage{amsmath} 
\usepackage{subfigure}
\usepackage{caption}
\usepackage{array}
\usepackage{multirow}
\usepackage{subcaption}
\usepackage{todonotes}
\usepackage{colortbl}
\usepackage{booktabs}
\usepackage{geometry}
\begin{document}

\title{An Open Source Enhancement Algorithm for PCB Defect Detection Based on YOLOv8 Model}
\title{VR-YOLO: Enhancing PCB Defect Detection with Viewpoint Robustness Based on YOLO\\
\thanks{\dag, These authors contributed equally to this work.\par $*$, Corresponding author: He Li.}}


\author{
    \IEEEauthorblockN{Hengyi Zhu$^{1\dag}$, Linye Wei$^{1\dag}$, He Li$^{2*}$}
    \IEEEauthorblockA{$^1$Chien-Shiung Wu College, Southeast University, Nanjing, China}
    \IEEEauthorblockA{$^2$School of Electronic Science \& Engineering, Southeast University, Nanjing, China}
    \IEEEauthorblockA{helix@seu.edu.cn}
}

\maketitle

\begin{abstract}
The integration of large-scale circuits and systems emphasizes the importance of automated defect detection of electronic components. The YOLO image detection model has been used to detect PCB defects and it has become a typical AI-assisted case of traditional industrial production. However, conventional detection algorithms have stringent requirements for the angle, orientation, and clarity of target images. 
In this paper, we propose an enhanced PCB defect detection algorithm, named VR-YOLO, based on the YOLOv8 model. This algorithm aims to improve the model's generalization performance and enhance viewpoint robustness in practical application scenarios. We first propose a diversified scene enhancement (DSE) method by expanding the PCB defect dataset by incorporating diverse scenarios and segmenting samples to improve target diversity. A novel key object focus (KOF) scheme is then presented by considering angular loss and introducing an additional attention mechanism to enhance fine-grained learning of small target features. Experimental results demonstrate that our improved PCB defect detection approach achieves a mean average precision (mAP) of 98.9\% for the original test images, and 94.7\% for the test images with viewpoint shifts (horizontal and vertical shear coefficients of ±0.06 and rotation angle of ±$10$ degrees), showing significant improvements compared to the baseline YOLO model with negligible additional computational cost.

\end{abstract}


\section{Introduction}
As the development of chiplet and 3D integrated circuit manufacturing technologies progresses, the complexity of circuit designs is continuously increasing\cite{yang2024challenges}. This escalation has made defect detection in chips, SoCs, or FPGA boards increasingly critical. Printed circuit boards (PCBs) are the essential interconnecting and structural foundation of integrated circuits, and their performance and reliability have long been the focus of research attention. Traditional manual inspection methods are gradually becoming obsolete due to their time-consuming nature, high labor intensity, and elevated false detection rates. In contrast, machine vision inspection technologies have gained prominence. By utilizing high-resolution imaging and advanced image processing algorithms, these technologies can rapidly and accurately identify and classify various surface defects\cite{liao2024chip}.

Due to the rapid progress of artificial intelligence, deep learning has demonstrated its vast potential in enhancing detection efficiency and accuracy within the chip manufacturing domain. In particular, the series of YOLO models and the use of OpenCV have empowered automated detection systems to identify various defects accurately and in real time, thereby ensuring product quality\cite{kieu2025deep}. Lightweight design like affinity propagation (AP) clustering algorithms for anchor boxes \cite{chen2024defect}, and model channel pruning \cite{meng2024real} have been proposed to enhance efficiency. Generative adversarial network (GAN) modules \cite{patel2024musap}, weighted bidirectional feature pyramid networks (BiFPN) \cite{bai2025improved}, and manual annotated low contrast defect detection methods have been utilized to improve defect detection accuracy.

Recent works, such as GS-YOLO \cite{zhu2025gs} and FDDC-YOLO~\cite{zheng2025fddc}, have made great progress in PCB defect detection by introducing new intersection over union (IoU) loss and attention mechanisms into YOLO models. However, their effectiveness in detecting minuscule defects in overlooked non-standard scenarios remains uncertain. These methods often fail to ensure accuracy and reliability under the varied viewpoints encountered in real-world applications. To address this, we propose VR-YOLO, designed specifically for PCB defect detection in diverse practical contexts, including shearing and rotation. With diversified scene enhancement and a novel object focus scheme, our YOLO augmentation across data and detection algorithms achieves remarkable success rates on multiple defect detection tasks.

The main contributions of the proposed methods can be summarized as follows:

\begin{itemize}
    \item[\textbf{\textbullet}] \textbf{Preprocessing} Our image augmentation simulates the complex practical PCB inspection environment, effectively coping with the viewpoint shifts of PCB images.
    
    \item[\textbf{\textbullet}] \textbf{Algorithm.} We propose an optimization algorithm based on the YOLO model, which significantly improves the accuracy of PCB defect detection with DSE and KOF.
    

    \item[\textbf{\textbullet}] \textbf{Achievements.} We achieve a mean average precision (mAP) of 98.9\% over original test images, while also realizing a mAP of 94.7\% over test images with simulated changes in viewpoint.

\end{itemize}

\section{Related work}

\subsection{Image Augmentation}
Cropping and rotating images are common methods to address the challenge of detecting small objects in object detection tasks. However, this practice diverges from the realities of actual image capture scenarios. Recently, efforts have been made to mitigate the negative impact of viewpoint shifts. In cancer detection\cite{mira2024early}, researchers have performed close-ups and multi-angle shots on tumors with prominent features. In sperm detection\cite{nawae2023comparative}, researchers have applied shearing and blurring to images of sperm under a microscope, in an attempt to mimic the microscope focusing at different distances and orientations.

\subsection{Loss Function}
The intersection over union (IoU) loss function is frequently used in image recognition and classification tasks. It measures the overlap between the predicted and true object region. Zheng et al.\cite{zheng2020distance} proposed DIoU (distance-IoU) to measure the distance, overlap, and shape between predicted and real bounding boxes. Based on this, they introduced CIoU (complete-IoU), which takes into account the aspect ratio of the bounding boxes, resulting in more stable regression. However, the loss will be zero if the aspect ratios of the predicted and real boxes are the same. Zhang et al.\cite{zhang2022focal} proposed EIoU (efficient-IoU), which separately considers the width and length of the border, addressing the issue of CIoU. Gevorgyan\cite{gevorgyan2022siou} further introduced SIoU (SCYLLA-IoU), which incorporates angle loss and quickly adjusts the predicted box to the nearest axis, reducing the model's degree of freedom.

\subsection{Attention Mechanism}
The attention mechanism (AM) is a technique that imitates human attention distribution to focus on important parts of input data. Xu et al.\cite{xu2015show} proposed a spatial attention mechanism, which adjusts the degree of attention based on the spatial location of input data. However, the pure spatial attention mechanism does not give sufficient consideration to global information. Hu et al.\cite{hu2018squeeze} introduced SENet (squeeze-and-excitation net), a channel attention mechanism that adjusts its attention based on the importance of different input data channels. However, channel attention mechanisms may prioritize channels with higher response and overlook crucial but low-response channels. CBAM (convolutional block attention module), an attention mechanism proposed by Woo et al.~\cite{woo2018cbam}, enhances the model's generalization performance by improving the feature map's representation ability through the combination of channel and spatial information.

\section{Data Processing and Algorithm Improvement}

\subsection{Diversified Scene Enhancement (DSE)}

\subsubsection{Scene Augmentation}

In this paper, we address the challenges associated with the strict requirements on the uniformity of shooting angle and direction for PCB defect detection in the current algorithms. These requirements significantly impact the accuracy of defect type recognition and border annotation, as even minor variations in the angle and direction of PCB images can lead to considerable inaccuracies. To mitigate these issues, we introduce an approach that involves applying shearing and rotation operations to the PCB images in our dataset. This method simulates images of the same PCB board captured from varying viewpoints.


To simulate different viewing angles, we perform a shearing operation on the PCB image using a two-dimensional affine transformation as described below:
\begin{equation}\label{eq}
P^{'}=M \cdot P
\end{equation}
\begin{equation}\label{eq}
P^{'}=\begin{bmatrix}
    	x^{'} \\
    	y^{'} \\
	1
\end{bmatrix}
P=\begin{bmatrix}
    	x \\
    	y \\
	1
\end{bmatrix}
M=\begin{bmatrix}
    	S_x & Sh_x & T_x \\
    	Sh_y & S_y & T_y \\
	0 & 0 & 1
\end{bmatrix}
\end{equation}
where \(P\) represents the homogeneous coordinates before the transformation, and \(P'\) represents the homogeneous coordinates after the transformation. \(M\) denotes the two-dimensional affine transformation matrix. \(S_x\) and \(S_y\) are the scaling factors in the horizontal and vertical directions, respectively. \(Sh_x\) and \(Sh_y\) represent the shearing factors in the horizontal and vertical directions, respectively. \(T_x\) and \(T_y\) are the translations in the horizontal and vertical directions, respectively. For shearing purposes, we set \(S_x\) and \(S_y\) to 1 and \(T_x\) and \(T_y\) to 0. The PCB image's width and height are normalized to 1, simplifying the affine transformation as follows:
\begin{equation}\label{eq}
\begin{bmatrix}
    	x^{'} \\
    	y^{'} 
\end{bmatrix}
=\begin{bmatrix}
    	1 & Sh_x \\
    	Sh_y & 1
\end{bmatrix}
\begin{bmatrix}
    	x \\
    	y
\end{bmatrix}
\end{equation}



For rotation operations, we utilize the following equation on the PCB image:
\begin{equation}\label{eq}
\begin{bmatrix}
    	x^{'} \\
    	y^{'} 
\end{bmatrix}
=\begin{bmatrix}
    	\cos(\theta) & -\sin(\theta) \\
    	\sin(\theta) & \cos(\theta)
\end{bmatrix}
\begin{bmatrix}
    	x \\
    	y
\end{bmatrix}
\end{equation}
where $\theta$ represents the image rotation angle and is positive counterclockwise.



\subsubsection{Sample Refinement}

In the data preprocessing phase, we initially execute a 2×2 segmenting operation on each original image, dividing it into four smaller segments. This procedure not only diminishes the image size during model training, thereby conserving computational resources, but also augments the model's ability to detect smaller-sized defects.

Moreover, we implement a blurring technique to emulate PCB images captured at varying focus distances. This is achieved by randomly applying a Gaussian blur with a radius ranging from 1 to 5 pixels. These enhancements significantly improve the dataset's capacity to mimic the diversity of image quality variations encountered in actual production settings. To accommodate the input requirements of the YOLOv8 detection model, all image segments are uniformly resized to a resolution of 640×640. This uniform scaling ensures that the model comprehensively learns the characteristics of various defects at a consistent size, thereby laying a solid foundation for subsequent training and detection efforts. The entire DSE workflow is illustrated in Fig. 1.
\begin{figure}[!tb]
\centerline{\includegraphics[width=0.45\textwidth]{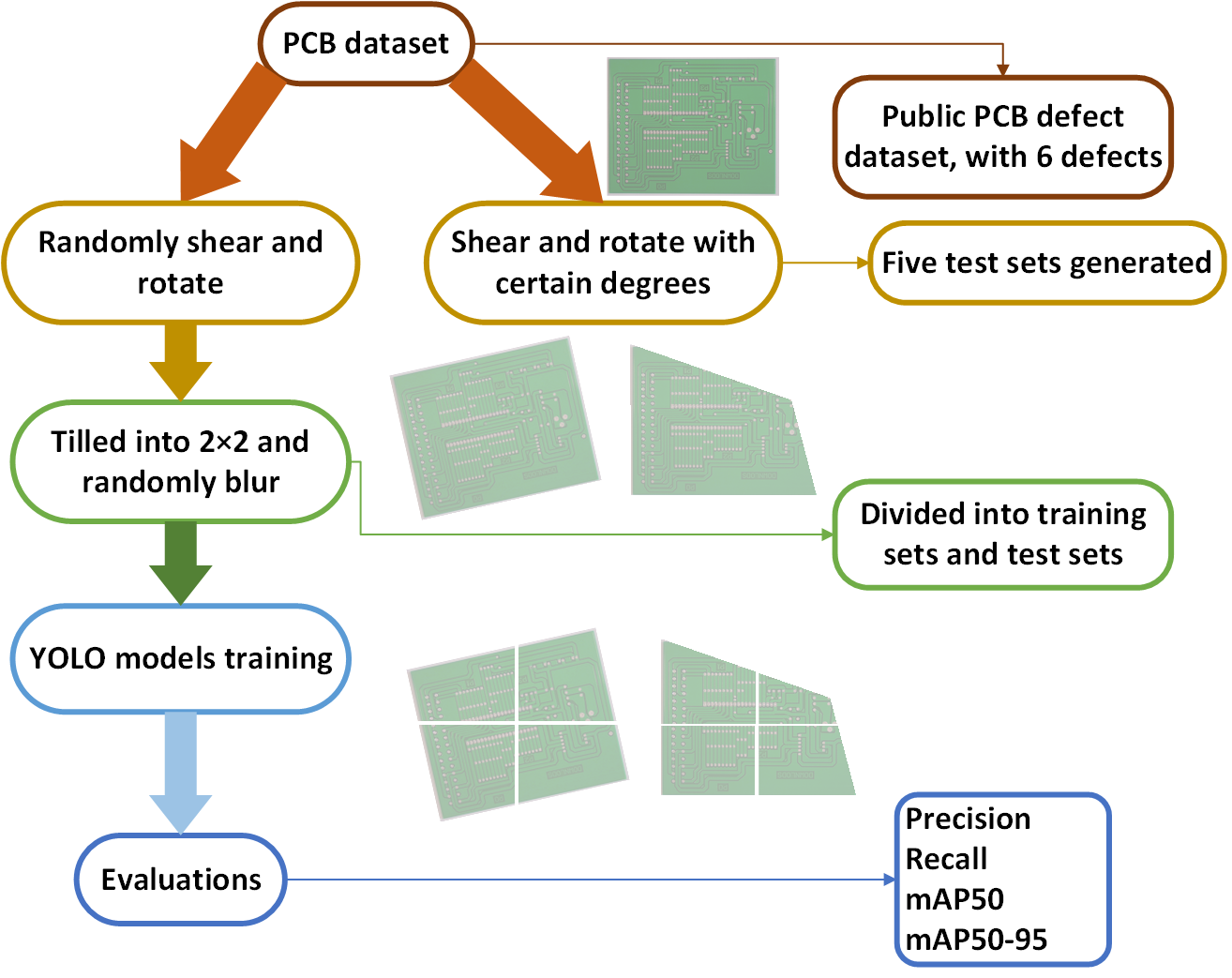}}
\caption{Complete flowchart of DSE process}
\end{figure}

\subsection{Key Object Focus (KOF)}

\subsubsection{Loss Function Optimization}

In YOLOv8, the evolution of loss functions from GIoU through DIoU to CIoU has significantly improved the prediction of bounding boxes by incorporating length, shape, and position. However, the directionality of bounding boxes has been overlooked, which is crucial for tasks such as PCB defect detection where image capture angles vary widely. This gap is addressed by transitioning to the SIoU loss function. SIoU is formulated to take into account angle, distance, shape, and base loss, which enhances the model's accuracy in scenarios where object orientation is a key factor. The SIoU loss function is defined as follows:

\begin{equation}\label{eq}
\begin{aligned}
L_{\text{SIoU}} &= 1 - I o U + \frac{\Delta + \Omega}{2} \\
&= 1 - \frac{\left|B \cap B^{G T}\right|}{\left|B \cup B^{G T}\right|} + \frac{1}{2} \\
&\quad \left\{\sum_{t=x, y}\left[1-e^{-(2-\Lambda)\rho_t}\right] + \sum_{t=w, h}\left(1-e^{-\omega_t}\right)^\theta\right\}
\end{aligned}
\end{equation}
\begin{equation}\label{eq}
\begin{split}
\Lambda=1-2 * \sin ^2\left(\arcsin \left(\frac{c_h}{\sigma}\right)-\frac{\pi}{4}\right) \\
\end{split}
\end{equation}
where $IoU$ represents the intersection ratio between the ground truth bounding box $B^{G T}$ and the predicted bounding box $B$, $\Delta$ represents the distance loss between the real frame and the predicted frame, $\Omega$ represents the shape loss between the real frame and the predicted frame, and $\Lambda$ represents the angle loss between the real frame and the predicted frame; \(c_h\), $\sigma$ are the height difference and distance between the center point of the real box and the prediction box, while $\rho_t$ ($t=x,y$) and $\omega_t$ ($t=w,h$) are the normalized squared differences in ($x, y$) and relative differences in (width, height) between them. By factoring in vector angles between expected regressions, SIoU redefines the angle penalty metric, enabling the predicted box to swiftly align with the nearest axis and effectively reducing the model's total degrees of freedom.

\subsubsection{CBAM Attention Mechanism}

In the realm of computer vision recognition utilizing deep learning, the attention mechanism employs a mask to create an additional layer of weights, marking crucial information within the image. This allows the neural network to focus on the key features of the recognition target during the learning process, thereby enhancing the effectiveness of the learning. Given the complexity of image details and the presence of minuscule defect targets in PCB defect detection, this study integrates the Convolutional Block Attention Module (CBAM) attention mechanism into the YOLOv8 network.
 
The CBAM attention mechanism integrates the channel attention module and the spatial attention module. Initially, it generates a channel attention map by leveraging the relationships among the feature channels, aiding in the processing of intricate texture information within PCB images. Subsequently, a spatial attention map is produced based on the spatial relationships between features. This is particularly advantageous for PCB defect detection, as it emphasizes local feature alterations, catering to the nuances of small defect targets and their tendency to frequently change locations within PCB datasets. Moreover, as a versatile and lightweight module, CBAM can be seamlessly integrated into neural networks, promising substantial enhancements in learning outcomes with minimal increases in computational demands. The architecture of the CBAM attention mechanism is depicted in Fig. 2.
\begin{figure}[!tb]
\centerline{\includegraphics[width=0.45\textwidth]{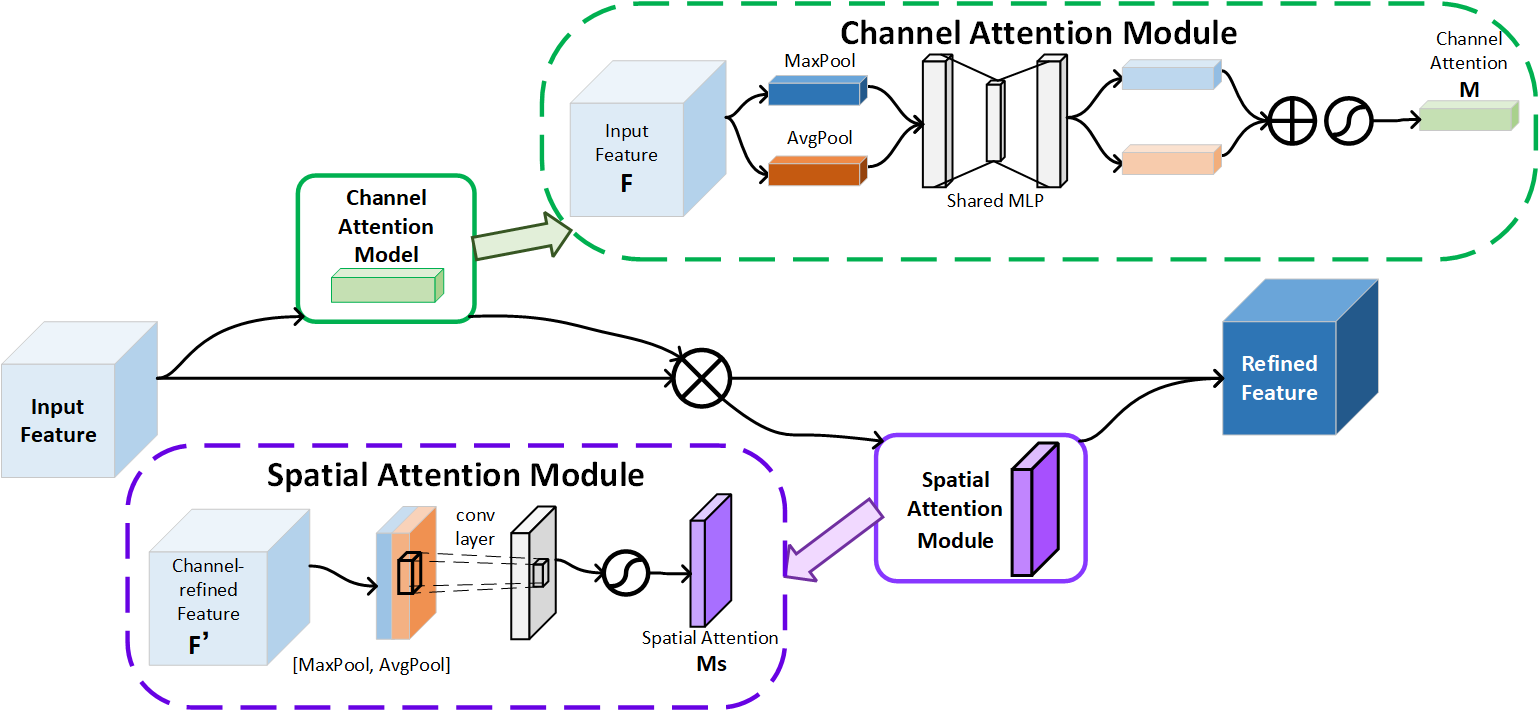}}
\caption{Network structure of CBAM attention mechanism}
\end{figure}

Following the incorporation of the CBAM attention mechanism, the modified backbone network structure of the YOLO model is illustrated in Fig. 3. 
\begin{figure}[!tb]
\centerline{\includegraphics[width=0.45\textwidth]{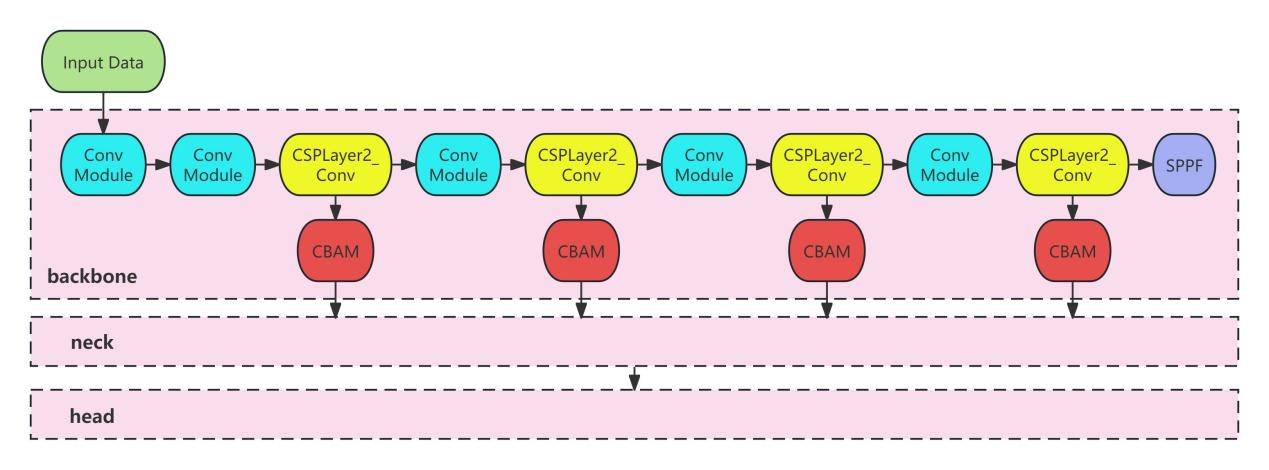}}
\caption{Backbone network structure of YOLOv8}
\end{figure}

\section{Experiments and Result Analysis}

\subsection{Experimental Environments}

\subsubsection{Original Dataset}
This research utilized the publicly available synthetic printed circuit board (PCB) defect dataset from the Peking University Open Laboratory of Intelligent Robotics as its foundation \cite{PKURoboticsDatasets}. The dataset comprises 693 images featuring six distinct types of defects (Missing\textunderscore{hole}, Mouse\textunderscore{bite}, Open\textunderscore{circuit}, Short, Spur, Spurious\textunderscore{copper}) suitable for tasks such as image detection, classification, and registration. The variety of defects present in the dataset is showcased in Fig. 4.
\begin{figure}[!tb]
    \centering    
	\subfigure{
        \includegraphics[width=0.145\textwidth]{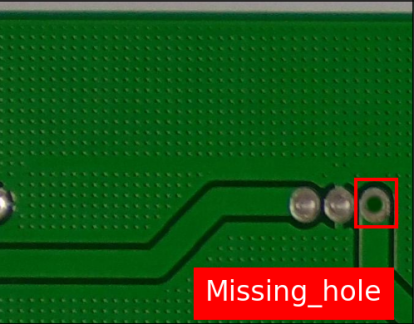}
    }
	\subfigure{
        \includegraphics[width=0.145\textwidth]{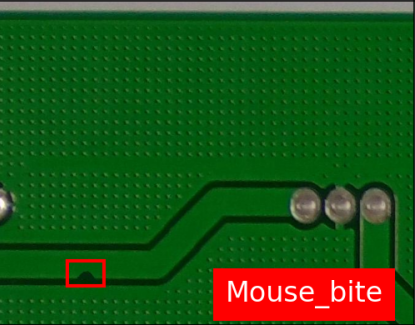}
    }
	\subfigure{
        \includegraphics[width=0.145\textwidth]{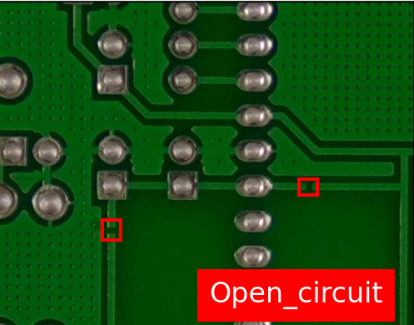}
    }
	\subfigure{
        \includegraphics[width=0.145\textwidth]{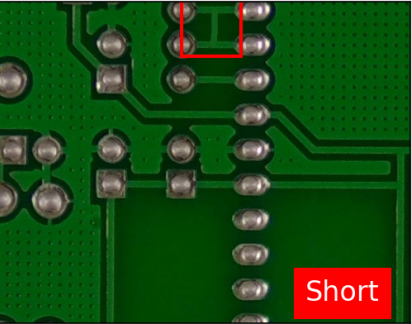}
    }
	\subfigure{
        \includegraphics[width=0.145\textwidth]{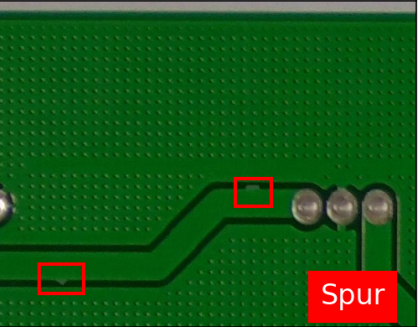}
    }
	\subfigure{
        \includegraphics[width=0.145\textwidth]{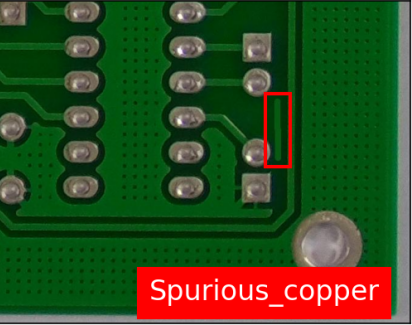}
    }
    \caption{Six types of defects in the dataset}
\end{figure}

The dataset was randomly partitioned into training and validation sets in an 80:20 split, respectively. Table I details the distribution of image types and defects within these sets.

\begin{table}[!tb]
\begin{center}
\caption{Types of images and defects in the original dateset}
\centering
\setlength{\tabcolsep}{8pt} 
\renewcommand{\arraystretch}{1.2}
\resizebox{\linewidth}{!}{
\begin{tabular}{c|cccc}
\toprule
\textbf{}&\multicolumn{2}{c}{\textbf{Train Set}} &\multicolumn{2}{c}{\textbf{Val Set}}\\
\midrule 
\textbf{Defect} 
&\textbf{\textit{Number of}}&\textbf{\textit{Number of}}&\textbf{\textit{Number of}}&\textbf{\textit{Number of}} \\
\textbf{type} 
&\textbf{\textit{images}}&\textbf{\textit{defects}}
&\textbf{\textit{images}}&\textbf{\textit{defects}} \\
\hline
Missing\textunderscore{hole}&92&398&23&99\\ 
Mouse\textunderscore{bite}&92&392&23&100\\
Open\textunderscore{circuit}&92&377&24&105\\
Short&92&382&24&109\\
Spur&92&391&23&97\\
Spurious\textunderscore{copper}&92&397&24&106\\
\hline
\textbf{Total}&552&2337&141&616\\
\bottomrule
\end{tabular}
}
\label{Table 1}
\end{center}
\end{table}

\subsubsection{Experimental Environment}
The experimental setup included hardware comprising two Intel(R) Xeon(R) Gold 6126 CPUs operating at 2.60GHz, with a system memory of 192GB. For graphical processing, it was equipped with 12 NVIDIA Tesla V100 GPUs, each boasting 32GB of video memory. Regarding the software environment, PyTorch version 2.1.2 was selected as the deep learning framework for this study.

\subsubsection{Experimental Dataset}

Building upon the original dataset, this study simulated viewpoint variations observed in actual PCB inspection scenarios by establishing a test dataset, PCB\textunderscore{DATASET\textunderscore{Contrast}}, and an enhanced dataset, PCB\textunderscore{DATASET\textunderscore{Shear\textunderscore{Rotate}}}, for experimental purposes. 

The PCB\textunderscore{DATASET\textunderscore{Contrast}} dataset was formed by applying either shearing or rotation operations to the test set. The newly generated images and corresponding label files replaced the original images and their labels.

The training dataset PCB\textunderscore{DATASET\textunderscore{Shear\textunderscore{Rotate}}} performs one shear and one rotation operation on all images in the original dataset. The generated images and corresponding label files are added to the dataset instead of replacing them, and the number of samples is expanded by 3 times. Then, the image is segmented and blurred to realize the scene expansion and feature generalization of the complete DSE process.




\subsection{Model Hyperparameter Settings}

During the training and testing phases, the image dimensions were standardized to 640×640 pixels, with the training process capped at a maximum of 1000 steps. The YOLOv8 model was employed, featuring a depth scaling factor of 0.33 and a width scaling factor of 0.25, while capping the maximum channel count at 1024. The CBAM attention mechanism was configured with three layers, having channel counts of 128, 256, and 512, respectively, and a window size of 7. 

\subsection{Evaluation Metrics}
The model's performance in object detection was assessed based on several key indicators: precision ($P$), recall ($R$) and mean average precision ($mAP$). The precision reflects the accuracy of the model's predictions, while the recall indicates the model's ability to detect all relevant objects. The $mAP$ (mean average precision) evaluates the precision of the model's detection capabilities. Detailed computation for these metrics has been provided within the available package \cite{yolov8_ultralytics}.

\subsection{Experimental Results and Analysis}

\subsubsection{Effect Analysis of Scene Augmentation}

In the initial phase of data processing, this study incorporates techniques such as image segmentation and Gaussian blurring. To simulate the viewpoint shift, both horizontal and vertical shear transformations with coefficients of ±0.06 and rotation angles of ±$10^\circ$ were applied. The outcomes were evaluated using precision ($P$) and recall ($R$), alongside the mean average precision ($mAP_{50}$) at a fixed threshold of 0.5, as summarized in Table II.

\begin{table}[!tb]
\begin{center}
\caption{Comparison of whether to perform scene augmentation or not}
\centering
\setlength{\tabcolsep}{8pt} 
\renewcommand{\arraystretch}{1.2}
\resizebox{\linewidth}{!}{
\begin{tabular}{c|ccc}
\toprule
Method & Precision & Recall & mAP50 \\
\midrule
Seg+Gauss & 93.6 & 92.7 & 94.7 \\
\hline
Non Seg+Gauss & 88.4 & 78.8 & 82.7 \\
\bottomrule
\end{tabular}
}
\label{Table 2}
\end{center}
\end{table}

The findings reveal that preprocessing techniques like image segmentation and Gaussian blurring can significantly enhance the model's detection accuracy across various metrics. These techniques significantly enhance the model's ability to detect defects in PCB images, especially in situations involving viewpoint shifts. 

\subsubsection{Comparative Analysis of Loss Function}

The YOLOv8 algorithm integrates various metrics for evaluating loss functions. By comparing the network's convergence rate using the model's default CIoU and the SIoU employed in this study, the variations in the bounding box loss (Box Loss) and class prediction loss (CLS Loss) relative to the number of training steps are depicted in Fig. 5.
\begin{figure}[!tb]
 \centerline{\includegraphics[width=0.48\textwidth]{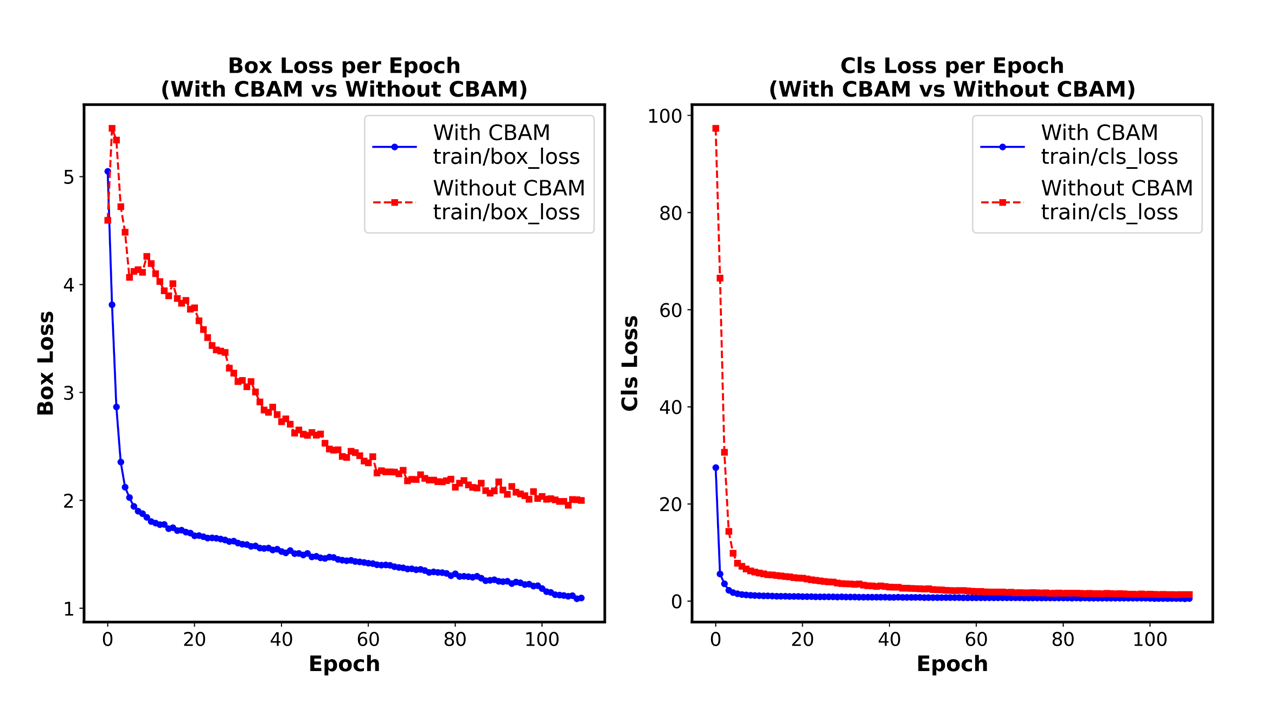}}

    \caption{Box loss and CLS loss}
\end{figure}

The experimental outcomes indicate that by adopting the SIoU loss function, our detection model exhibits a marked improvement in reducing both the bounding box loss and the class prediction loss. This leads to a faster convergence of the model, enhancing its overall performance and training efficiency.

\subsubsection{Analysis of Attention Mechanism}

To improve the original YOLOv8 visual recognition algorithm’s capability for defect detection, this study integrates the CBAM attention mechanism, aiming to enhance performance in both channel and spatial dimensions. The effectiveness of incorporating the CBAM attention mechanism was evaluated by plotting the recall curve and precision curve against the Intersection over Union (IoU) threshold, as depicted in Fig. 6.
\begin{figure}[!tb]
    \centerline{\includegraphics[width=0.48\textwidth]{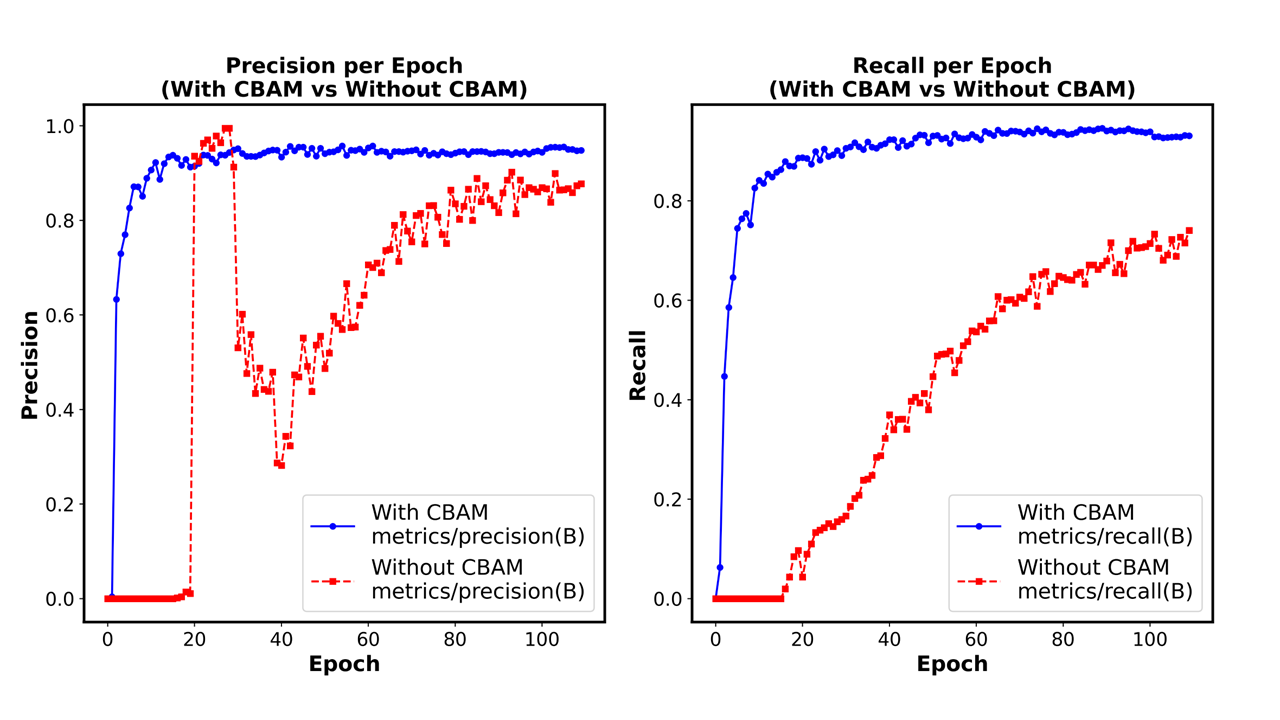}}
    \caption{Precision and recall curves}
\end{figure}

The experimental findings underscore a significant enhancement in the recall rate at higher IoU thresholds after integrating the CBAM attention mechanism. This improvement maintains a high recall rate over a broader threshold range, indicating superior model performance expectations. 

\subsubsection{PCB Defect Detection From Different Viewpoints}

This research establishes four distinct PCB\textunderscore{DATASET\textunderscore{Contrast}} test datasets to examine the model's capability in identifying PCB defects from varying viewpoints. These datasets correspond to manipulations such as Shear000+Rotate00, Shear003+Rotate05 (±0.03 shear coefficients, ±$5^\circ$ rotation), Shear006+Rotate05 (±0.06 shear coefficients, ±$5^\circ$ rotation), and Shear006+Rotate10 (±0.06 shear coefficients, ±$10^\circ$ rotation).

The performance of the original YOLOv8 model and the enhanced model proposed in this paper were analyzed across these datasets. Compared to recent advancements such as FDDC-YOLO\cite{zheng2025fddc}, CDS-YOLO\cite{shao2024enhanced}, DsP-YOLO\cite{zhang2024dsp}, Inner-SIoU\cite{du2024lightweight} and SSHP-YOLO\cite{wang2025sshp}, our model demonstrates the best performance in terms of precision, recall and mAP50 on raw PCB images as summarized in Table III.


\begin{table}[!tb]
\begin{center}
\caption{Comparison of raw PCB image test}
\centering
\setlength{\tabcolsep}{8pt} 
\renewcommand{\arraystretch}{1.2}
\resizebox{\linewidth}{!}{
\begin{tabular}{c|ccc}
\toprule
Method & Precision & Recall & mAP50 \\
\midrule
FDDC-YOLO \cite{zheng2025fddc} & 85.3 & 82.6 & 83.6 \\
\hline
CDS-YOLO \cite{shao2024enhanced} & 97.4 & 92.9 & 95.3 \\
\hline
DsP-YOLO \cite{zhang2024dsp} & 94.7 & 94.3 & 95.8 \\
\hline
Inner-SIoU \cite{du2024lightweight} & 96.6 & 92.1 & 96.8 \\
\hline
SSHP-YOLO \cite{wang2025sshp} & 96.9 & 96.2 & 97.8 \\
\hline
\rowcolor{gray!20}
\textbf{VR-YOLO} & \textbf{97.9} & \textbf{96.8} & \textbf{98.9} \\
\bottomrule
\end{tabular}
}
\label{Table 4}
\end{center}
\end{table}

Moreover, the VR-YOLO proposed in this study has an effect on the viewpoint shift of PCB images. Comparative analysis of the original and enhanced models across metrics like Precision, Recall, $mAP_{50}$, and $mAP_{50-95}$(average value of $mAP$ for $IoU$ thresholds ranging from 0.5 to 0.95 with a step size of 0.05) is depicted in Fig. 7, highlighting the enhanced model's superior performance.

\begin{figure}[!tb]
    \centering    
	\subfigure{
        \includegraphics[width=0.48\textwidth]{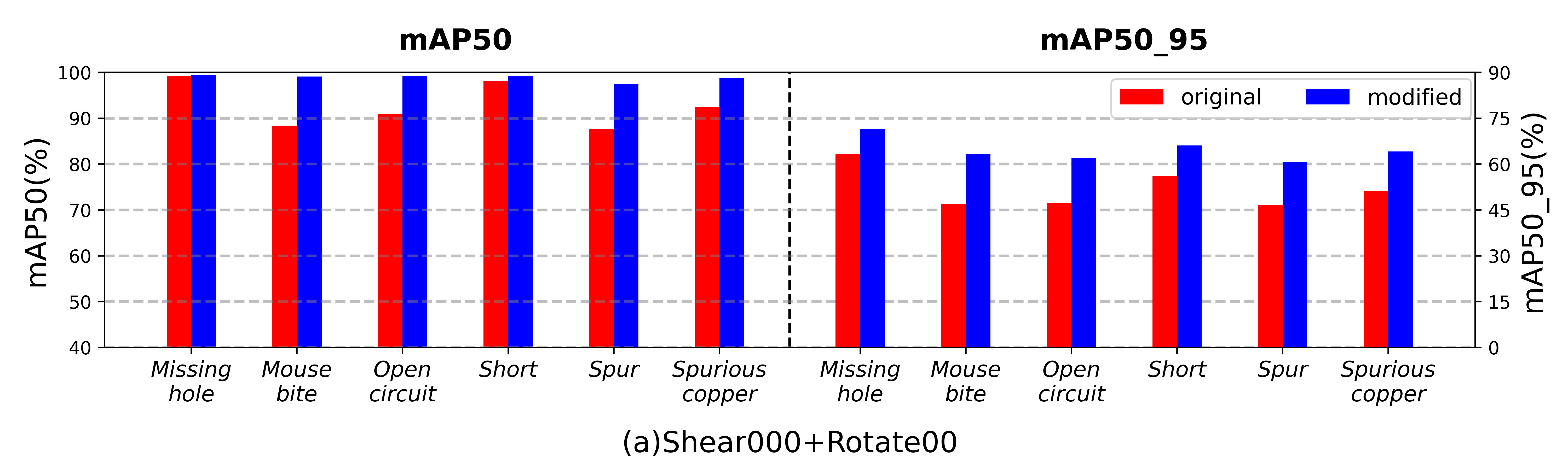}
        \label{fig:$Shear000+Rotate00$}
    }
	\subfigure{
        \includegraphics[width=0.48\textwidth]{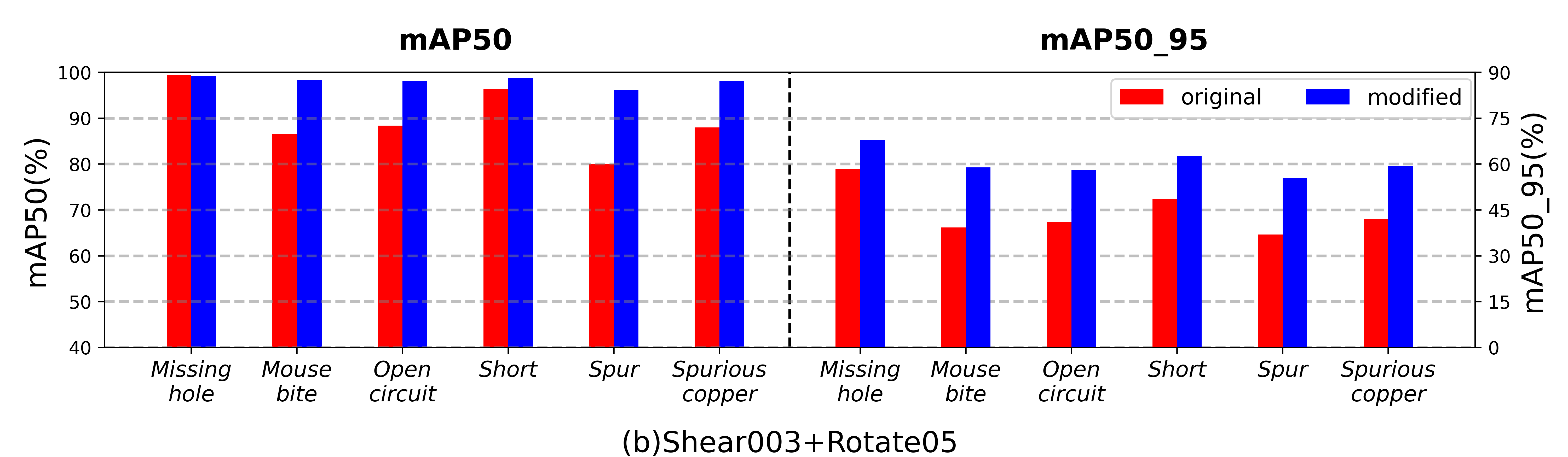}
        \label{fig:$Shear003+Rotate05$}
    }
	\subfigure{
        \includegraphics[width=0.48\textwidth]{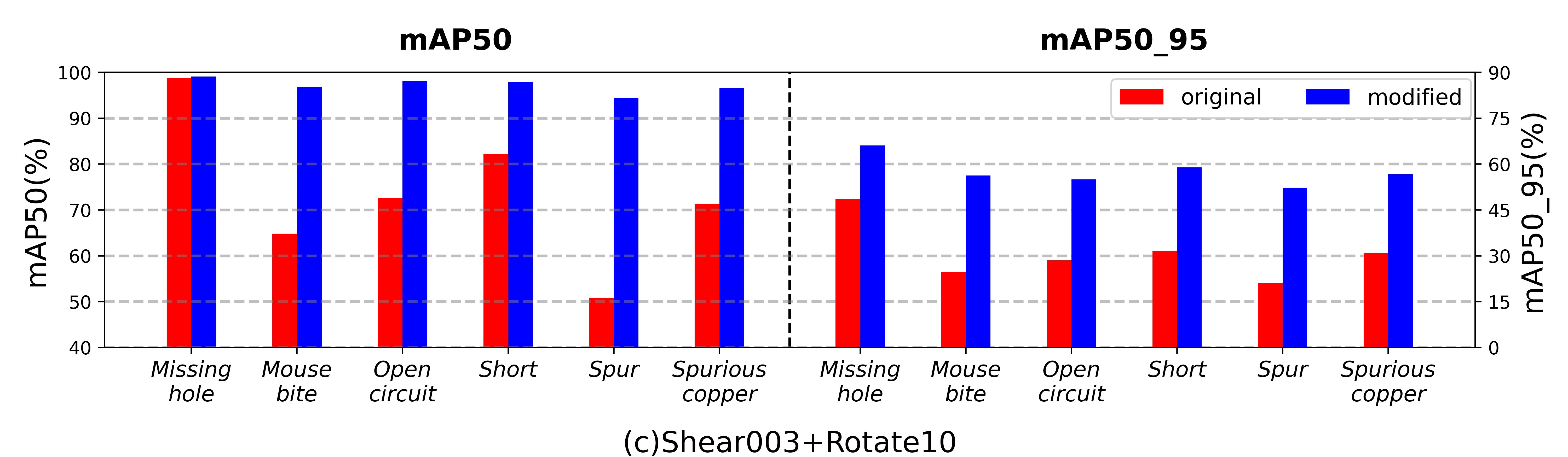}
        \label{fig:$Shear003+Rotate10$}
    }
	\subfigure{
        \includegraphics[width=0.48\textwidth]{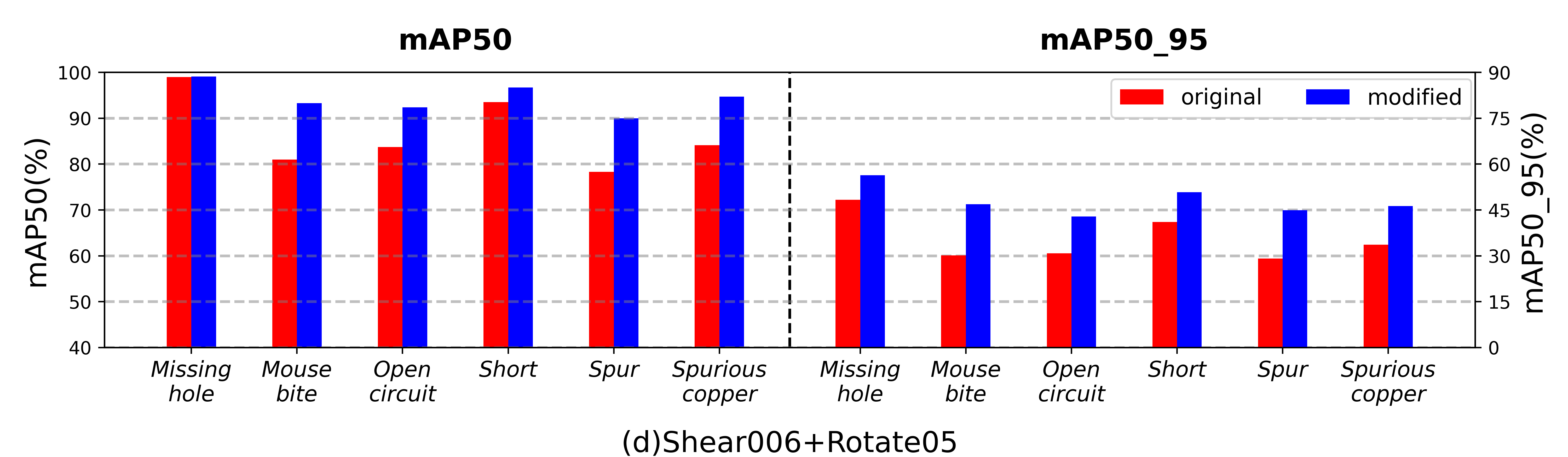}
        \label{fig:$Shear006+Rotate05$}
    }
    \subfigure{
        \includegraphics[width=0.48\textwidth]{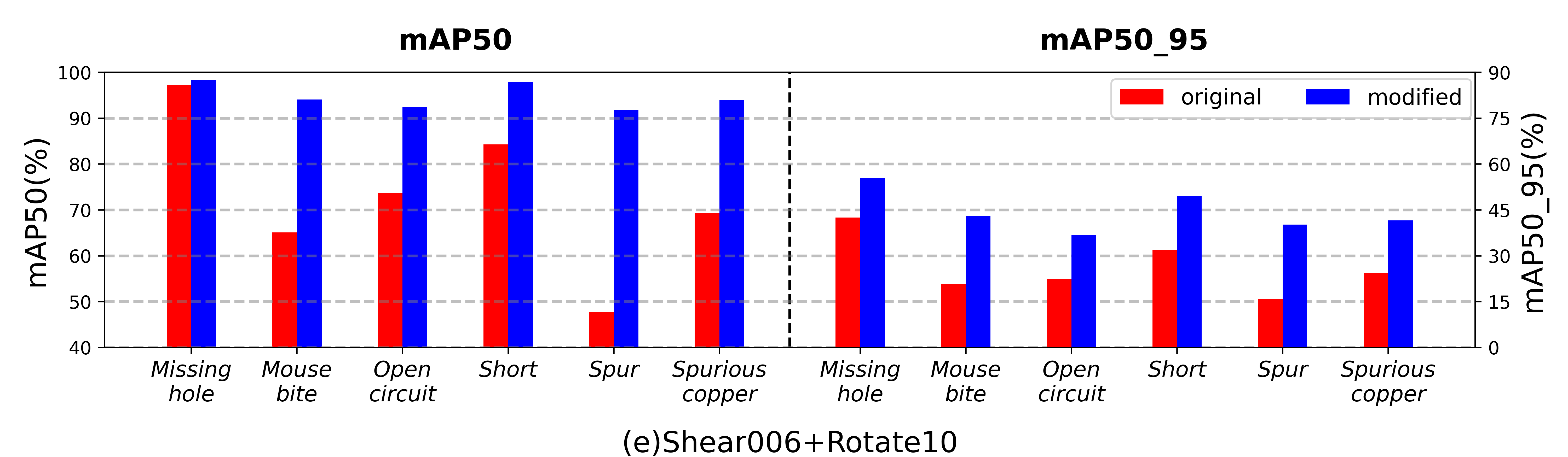}
        \label{fig:$Shear006+Rotate10$}
        }
    \caption{Comparison of mAP50 and mAP50-95 under the two models}
    
\end{figure}

The findings affirm that VR-YOLO proficiently addresses challenges posed by directional and angular deviations in PCB images in real-world scenarios, marking a notable improvement in accuracy with viewpoint robustness. Notably, our proposed enhanced PCB defect detection method has little impact on computational efficiency. 

\section{Conclusion}
This paper improves the robustness and accuracy of the PCB defect detection algorithm for practical application scenarios. By employing diversified scene enhancement and key object focus measures, we improve the model's convergence speed, enhance the algorithm's generalization performance for detecting small target feature defects, and effectively address the high requirements of traditional detection algorithms regarding image angle, direction, and resolution. Future research could focus on optimizing the model's structure to better adapt to complex production environments, considering the practical deployment and application of the algorithms.

\section*{Acknowledgment}
%
This work is supported in part by the National Natural Science Foundation of China under Grant 62304037, in part by the Natural Science Foundation of Jiangsu Province under Grant BK20230828, in part by the Young Elite Scientists Sponsorship Program by CAST under Grant 2022QNRC001, in part by the Southeast University Interdisciplinary Research Program for Young Scholars under Grant 2024FGC1005, in part by the Fundamental Research Funds for the Central Universities under Grant 2242025K30008, and in part by the Start-up Research Fund of Southeast University under Grant RF1028623173.

\bibliographystyle{IEEEtran}
\bibliography{reference}

\begin{thebibliography}{10}
\providecommand{\url}[1]{#1}
\csname url@samestyle\endcsname
\providecommand{\newblock}{\relax}
\providecommand{\bibinfo}[2]{#2}
\providecommand{\BIBentrySTDinterwordspacing}{\spaceskip=0pt\relax}
\providecommand{\BIBentryALTinterwordstretchfactor}{4}
\providecommand{\BIBentryALTinterwordspacing}{\spaceskip=\fontdimen2\font plus
\BIBentryALTinterwordstretchfactor\fontdimen3\font minus \fontdimen4\font\relax}
\providecommand{\BIBforeignlanguage}[2]{{%
\expandafter\ifx\csname l@#1\endcsname\relax
\typeout{** WARNING: IEEEtran.bst: No hyphenation pattern has been}%
\typeout{** loaded for the language `#1'. Using the pattern for}%
\typeout{** the default language instead.}%
\else
\language=\csname l@#1\endcsname
\fi
#2}}
\providecommand{\BIBdecl}{\relax}
\BIBdecl

\bibitem{yang2024challenges}
Z.~Yang, S.~Ji, X.~Chen, J.~Zhuang, W.~Zhang, D.~Jani, and P.~Zhou, ``Challenges and opportunities to enable large-scale computing via heterogeneous chiplets,'' in \emph{2024 29th Asia and South Pacific Design Automation Conference (ASP-DAC)}.\hskip 1em plus 0.5em minus 0.4em\relax IEEE, 2024, pp. 765--770.

\bibitem{liao2024chip}
C.~Liao, C.~Zhan, W.~He, H.~Gao, and P.~Zhou, ``Chip surface character detection system based on machine vision,'' in \emph{2024 4th International Conference on Computer Science, Electronic Information Engineering and Intelligent Control Technology (CEI)}.\hskip 1em plus 0.5em minus 0.4em\relax IEEE, 2024, pp. 413--418.

\bibitem{kieu2025deep}
X.-T. Kieu, V.-T. Nguyen, D.-T. Chu, X.-H. Van, M.~Van, S.-F. Su, and X.-T. Phan, ``Deep learning-enhanced defects detection for printed circuit boards,'' \emph{Results in Engineering}, p. 104067, 2025.

\bibitem{chen2024defect}
W.~Chen, H.~Zhao, and Z.~Wang, ``Defect detection model of printed circuit board components based on the fusion of multi-scale features and efficient channel attention mechanism,'' \emph{IEEE Access}, 2024.

\bibitem{meng2024real}
D.~Meng, X.~Xu, Z.~Jiang, and L.~Xu, ``Real-time detection of insulator defects with channel pruning and channel distillation,'' \emph{Applied Sciences}, vol.~14, no.~19, p. 8587, 2024.

\bibitem{patel2024musap}
N.~Patel, ``{{MuSAP-GAN}}: printed circuit board defect detection using multi-level attention-based printed circuit board with generative adversarial network,'' \emph{Electrical Engineering}, pp. 1--18, 2024.

\bibitem{bai2025improved}
L.~Bai and W.~H. Xu, ``Improved printed circuit board defect detection scheme,'' \emph{Scientific Reports}, vol.~15, no.~1, p. 2389, 2025.

\bibitem{zhu2025gs}
H.~Zhu, L.~Dong, H.~Ren, H.~Zhuang, and H.~Li, ``{{GS-YOLO}}: A lightweight identification model for precision parts,'' \emph{Symmetry}, vol.~17, no.~2, p. 268, 2025.

\bibitem{zheng2025fddc}
H.~Zheng, J.~Peng, X.~Yu, M.~Wu, Q.~Huang, and L.~Chen, ``{{FDDC-YOLO}}: an efficient detection algorithm for dense small-target solder joint defects in {{PCB}} inspection,'' \emph{Journal of Real-Time Image Processing}, vol.~22, no.~2, p.~83, 2025.

\bibitem{mira2024early}
E.~S. Mira, A.~M. Saaduddin~Sapri, R.~F. Aljehan{\i}, B.~S. Jamb{\i}, T.~Bashir, E.-S.~M. El-Kenawy, and M.~Saber, ``Early diagnosis of oral cancer using image processing and artificial intelligence.'' \emph{Fusion: Practice \& Applications}, vol.~14, no.~1, 2024.

\bibitem{nawae2023comparative}
M.~Nawae, P.~Maneelert, C.~Choksuchat, T.~Phairatana, and J.~Jaruenpunyasak, ``A comparative study of {{YOLO}} models for sperm and impurity detection based on proposed augmentation in small dataset,'' in \emph{2023 15th International Conference on Information Technology and Electrical Engineering (ICITEE)}.\hskip 1em plus 0.5em minus 0.4em\relax IEEE, 2023, pp. 305--310.

\bibitem{zheng2020distance}
Z.~Zheng, P.~Wang, W.~Liu, J.~Li, R.~Ye, and D.~Ren, ``{{Distance-IoU}} loss: Faster and better learning for bounding box regression,'' in \emph{Proceedings of the AAAI conference on artificial intelligence}, vol.~34, no.~07, 2020, pp. 12\,993--13\,000.

\bibitem{zhang2022focal}
Y.-F. Zhang, W.~Ren, Z.~Zhang, Z.~Jia, L.~Wang, and T.~Tan, ``Focal and efficient iou loss for accurate bounding box regression,'' \emph{Neurocomputing}, vol. 506, pp. 146--157, 2022.

\bibitem{gevorgyan2022siou}
Z.~Gevorgyan, ``{{SIoU}} loss: More powerful learning for bounding box regression,'' \emph{arXiv preprint arXiv:2205.12740}, 2022.

\bibitem{xu2015show}
K.~Xu, J.~Ba, R.~Kiros, K.~Cho, A.~Courville, R.~Salakhudinov, R.~Zemel, and Y.~Bengio, ``Show, attend and tell: Neural image caption generation with visual attention,'' in \emph{International conference on machine learning}.\hskip 1em plus 0.5em minus 0.4em\relax PMLR, 2015, pp. 2048--2057.

\bibitem{hu2018squeeze}
J.~Hu, L.~Shen, and G.~Sun, ``Squeeze-and-excitation networks,'' in \emph{Proceedings of the IEEE conference on computer vision and pattern recognition}, 2018, pp. 7132--7141.

\bibitem{woo2018cbam}
S.~Woo, J.~Park, J.-Y. Lee, and I.~S. Kweon, ``Cbam: Convolutional block attention module,'' in \emph{Proceedings of the European conference on computer vision (ECCV)}, 2018, pp. 3--19.

\bibitem{PKURoboticsDatasets}
\BIBentryALTinterwordspacing
P.-H. Lab. (2022) Pku-market-phonepku-market-pcb. [Online]. Available: \url{https://robotics.pkusz.edu.cn/resources/dataset/}
\BIBentrySTDinterwordspacing

\bibitem{yolov8_ultralytics}
\BIBentryALTinterwordspacing
G.~Jocher, A.~Chaurasia, and J.~Qiu, ``Ultralytics yolov8,'' 2023. [Online]. Available: \url{https://github.com/ultralytics/ultralytics}
\BIBentrySTDinterwordspacing

\bibitem{shao2024enhanced}
M.~Shao, L.~Min, M.~Liu, X.~Li, X.~Li \emph{et~al.}, ``An enhanced network model for {{PCB}} defect detection: {{CDS-YOLO}},'' \emph{Journal of Real-Time Image Processing}, vol.~21, no.~6, pp. 1--14, 2024.

\bibitem{zhang2024dsp}
Y.~Zhang, H.~Zhang, Q.~Huang, Y.~Han, and M.~Zhao, ``{{DsP-YOLO}}: An anchor-free network with {{DsPAN}} for small object detection of multiscale defects,'' \emph{Expert Systems with Applications}, vol. 241, p. 122669, 2024.

\bibitem{du2024lightweight}
P.~Du and X.~Song, ``Lightweight target detection: An improved {{YOLO}}v8 for small target defect detection on printed circuit boards,'' in \emph{Proceedings of the 2024 International Conference on Generative Artificial Intelligence and Information Security}, 2024, pp. 329--334.

\bibitem{wang2025sshp}
J.~Wang, L.~Ma, Z.~Li, Y.~Cao, and H.~Zhang, ``{{SSHP-YOLO}}: A high precision printed circuit board ({{PCB}}) defect detection algorithm with a small sample,'' \emph{Electronics}, vol.~14, no.~2, p. 217, 2025.

\end{thebibliography}
\vspace{12pt}
\end{document}